\documentclass[conference]{IEEEtran}
\def\IEEEtitletopspace{18pt}
\IEEEoverridecommandlockouts    

\usepackage{multirow}
\usepackage{lipsum}
\usepackage{amsmath}
\usepackage{amssymb}
\usepackage{float}
\usepackage{stfloats}
\usepackage[letterpaper, top=0.75in, bottom=1in, left=0.75in, right=0.75in]{geometry}
\usepackage[ruled,vlined]{algorithm2e}
\usepackage{graphicx}
\usepackage{xcolor}
\graphicspath{{./Figures/}}

\title{Safety-Oriented Evaluation of Language Understanding Systems for Air Traffic Control}

\author{
\IEEEauthorblockN{Yujing Chang$^{1,*}$, Yash Guleria$^{2}$, Duc-Thinh Pham$^{1,3}$, Nhut-Huy Pham$^{1}$, Ningli Wang$^{1}$, Vu N. Duong$^{3}$, Sameer Alam$^{1}$}
\IEEEauthorblockA{$^{1}$ATMRI, Nanyang Technological University (NTU), Singapore}
\IEEEauthorblockA{$^{2}$School of Management, Indian Institute of Technology Mandi, India}
\IEEEauthorblockA{$^{3}$Centre of AI Research, VinUniversity, Vietnam}
\IEEEauthorblockA{$^{*}$Corresponding author}
}

\begin{document}
	
	\maketitle

	\begin{abstract}

        Air Traffic Control (ATC) is a safety-critical domain in which incorrect interpretation of instructions may lead to severe operational consequences. 
        While large language models (LLMs) demonstrate strong general performance, their reliability in operational ATC environments remains unclear. 
        Existing evaluation approaches, largely based on aggregate metrics such as F1 or macro accuracy, treat all errors uniformly and fail to account for the asymmetric consequences of high-risk semantic mistakes (e.g., incorrect runway identifiers or movement constraints).
        To address this gap, we propose a safety-oriented, consequence-aware evaluation framework tailored to ATC operations. Our results reveal that while current LLMs achieve reasonable aggregate accuracy, their operational reliability is severely limited. Evaluated on clean transcripts, the peak Risk Score reaches only 0.69, with most models scoring below 0.6 despite high macro-F1 performance.
        Further analysis shows that errors concentrate in high-impact entities despite relatively stable action-type classification, indicating structural grounding deficiencies.
        These findings highlight the necessity of consequence-aware evaluation protocols for the responsible deployment of AI-assisted ATC systems.

	\end{abstract}

     \begin{figure*}[!b]
    \centering
    \includegraphics[width=\textwidth]{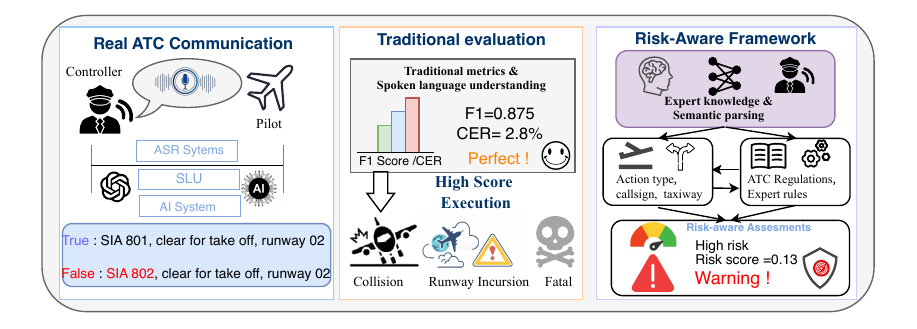}
    \caption{Overview of the evaluation pipeline, contrasting conventional semantic metrics with the proposed consequence-aware evaluation framework for ATC language understanding.}
    \label{fig:main}
    \end{figure*}

	\thispagestyle{empty}
	\pagestyle{empty}

\section{Introduction}
\label{sec:introduction_literature}

Air Traffic Control (ATC) is a core component of the global air transportation system, ensuring the safe and orderly movement of aircraft on runways, taxiways, and in controlled airspace. Misinterpretation of operational ATC instructions can lead to safety incidents, including runway incursions and mid-air collisions. These events represent some of the most critical safety hazards in aviation. Given the compact and context-dependent phraseology of ATC communications, accurate semantic understanding is essential for maintaining operational safety.

Recent advances in large language models (LLMs) and automatic speech recognition (ASR) systems have motivated growing interest in AI-assisted air traffic management \cite{openai2023gpt4, radford2023whisper}. Spoken language understanding (SLU) research has improved robustness to ASR errors through approaches such as contrastive learning, mixture-of-experts architectures, and speech-text alignment \cite{cheng2023c2a_slu, cheng2024moe_slu, dong2023cifpt, chen2023salm}, while ATC-specific studies have benefited from domain corpora, ASR benchmarks, transfer learning, and retrieval-enhanced scenario modeling for structured semantic extraction and operationally realistic evaluation \cite{hofbauer-etal-2008-atcosim, zuluagagomez2020automaticspeechrecognitionbenchmark, prabhavalkar2021atc_asr, wee2025adaptingautomaticspeechrecognition, thai2025speechtoroute, guleria2026text2traffic}.

Despite this progress, evaluation in both SLU and ATC remains centered on surface-level metrics such as word error rate (WER) \cite{graves2014speech}, character error rate (CER) \cite{graves2006ctc}, intent accuracy, and slot-level F1 \cite{mani-etal-2020-towards, arxiv2003.07692, yin-etal-2025-eclm}, which do not capture the operational consequences of semantic misunderstandings. This limitation is particularly acute in safety-critical settings, where misrecognizing a callsign in a high-risk taxi clearance carries far greater consequences than a minor slot omission in a routine exchange. Broader LLM safety research has similarly highlighted the heterogeneity of error severity \cite{xu2022sescore} and the need for behavioral evaluation beyond aggregate metrics \cite{ribeiro2020checklist}; related safety-critical domains such as medical AI have begun to adopt risk-stratified evaluation schemes \cite{guan2025radiologyrisk}. Nonetheless, systematic consequence-aware evaluation for real-time operational language understanding in ATC remains largely unexplored.

To address this gap, we propose a safety-oriented, risk-aware evaluation framework tailored to real-world ATC operations, as illustrated in Figure~\ref{fig:main}. The framework models structured semantic actions, including action types and safety-critical slots, and associates them with operational risk levels derived from expert knowledge and regulations, enabling consequence-aware scoring beyond aggregate accuracy. Using authentic ATC communication data, we systematically evaluate LLMs under both text-based and end-to-end voice-to-semantics settings. Through quantitative analysis and case studies, we reveal high-risk failure modes that remain hidden under conventional evaluation and show that, while current models achieve reasonable overall performance, reliability degrades significantly in safety-critical scenarios.

The main contributions of this work are:
\begin{itemize}
\item A safety-oriented, risk-aware evaluation framework for ATC language understanding, explicitly modeling safety-critical semantic actions and associated risk levels.
\item A systematic evaluation of LLMs on authentic ATC communications, including end-to-end voice-to-semantics assessment.
\item Quantitative and qualitative analyses revealing safety-critical failure modes not captured by conventional evaluation metrics.
\end{itemize}

	\section{Dataset}
	\label{sec:dataset}
	
	\subsection{Data Source and Initial Processing}

	The dataset used in this study builds upon real-world ATC communications collected at Singapore Changi Airport (ICAO: WSSS), originally introduced in \cite{thai2025speechtoroute}. Controller--pilot transmissions were recorded between 17--23 March 2025 across three ground-control frequencies (124.300 MHz, 121.850 MHz, 121.725 MHz) over two daily time windows. 

	The corpus focuses on surface movement instructions in ground control operations. To construct the evaluation subset, we extracted a continuous high-density traffic segment, removed utterances without identifiable aircraft callsigns, and reviewed the remaining data with a licensed controller and a commercial pilot. The final subset preserves natural operational sequences while providing approximately 1.2 hours of communication and around 1,000 annotated utterances.

\begin{table}[t]
\centering
\caption{Dataset statistics.}
\label{tab:dataset-stats}
\renewcommand{\arraystretch}{1.05}
\setlength{\tabcolsep}{4pt}
\footnotesize

\begin{minipage}[t]{0.49\columnwidth}
\centering
\textbf{(a) Utterance-level statistics}

\vspace{0.3\baselineskip}

\begin{tabular}{l r}
\hline
\textbf{Category} & \textbf{Count} \\
\hline
\multicolumn{2}{l}{\textit{By speaker}} \\
Pilot & 560 \\
Controller & 440 \\
\hline
\multicolumn{2}{l}{\textit{By intent}} \\
Instruction & 404 \\
Readback & 359 \\
Inform & 189 \\
Greet & 48 \\
\hline
\multicolumn{2}{l}{\textit{By action type}} \\
TAXI & 341 \\
INFORM & 169 \\
CONTACT & 149 \\
HOLD & 115 \\
PUSHBACK & 112 \\
GREET & 54 \\
GIVE\_WAY & 24 \\
UNKNOWN & 20 \\
STANDBY & 16 \\
\hline
\end{tabular}

\end{minipage}
\hfill
\begin{minipage}[t]{0.49\columnwidth}
\centering
\textbf{(b) Entity and risk statistics}

\vspace{0.3\baselineskip}

\begin{tabular}{l r}
\hline
\textbf{Entity} & \textbf{Count (\%)} \\
\hline
Total & 7,624 \\
O (outside) & 2,845 \\
non-O (entity) & 4,779 \\
\hline
Condition & 1,277 (26.7\%) \\
Taxiway & 1,114 (23.3\%) \\
Callsign & 988 (20.7\%) \\
Controller & 357 (7.5\%) \\
Greet & 270 (5.6\%) \\
Gate & 189 (4.0\%) \\
Frequency & 158 (3.3\%) \\
Report & 157 (3.3\%) \\
Qualifier & 124 (2.6\%) \\
Vehicle & 77 (1.6\%) \\
Runway & 68 (1.4\%) \\
\hline
\multicolumn{2}{l}{\textit{By action risk}} \\
High & 480 \\
Medium & 261 \\
Low & 259 \\
\hline
\end{tabular}

\end{minipage}

\end{table}

	\subsection{Risk-Oriented Semantic Reconstruction}

	To support risk-aware evaluation, each utterance was restructured into a unified action-level representation.

	\paragraph{Risk-informed schema.}
	A structured expert survey with five licensed tower controllers from Beijing Capital Airport Tower was conducted to assess the operational relevance of semantic components in ground-control communications. Their input informed the risk-oriented schema design, including the relative safety importance assigned to different entity categories.

	\paragraph{Action taxonomy.}
	Each utterance was mapped to one of nine canonical action types (HOLD, TAXI, GIVE\_WAY, CONTACT, PUSHBACK, INFORM, GREET, STANDBY, UNKNOWN), which were further stratified into three operational risk levels (HIGH, MEDIUM, LOW) according to their potential safety consequences in ground operations.

	\paragraph{Critical slots.}
	For each action type, a predefined set of critical slots was specified (e.g., callsign and boundary for HOLD; callsign and frequency for CONTACT). The dataset was annotated accordingly, producing structured representations containing action type, normalized slots, and associated risk level.

	This reconstruction converts free-form ATC utterances into risk-aware semantic units suitable for risk-aware evaluation.

	\subsection{Data analysis} 
	Table~\ref{tab:dataset-stats} summarizes the curated evaluation subset. Here, O denotes tokens outside annotated entity spans, while non-O denotes tokens assigned to semantic entity labels. The corpus shows balanced pilot--controller interaction, with instructions and readbacks comprising most exchanges, and maintains a stratified risk composition (48\% high, 26\% medium, 26\% low). Overall, it provides realistic surface-traffic dynamics with sufficient semantic and risk diversity for consequence-aware evaluation.


\section{Evaluation Metrics}
\label{sec:metrics}
As illustrated in Figure~\ref{fig:main}, treating all semantic errors uniformly fails to capture the strict safety risks inherent to ATC communications. To ensure operational grounding, our evaluation incorporates ATC regulations together with expert input grounded in licensed controllers' assessment of real-world operational risk (Table~\ref{tab:weight-survey}). This expert-informed process guided the action taxonomy, risk stratification, and consequence-aware weighting used throughout the evaluation. Under established ATC surface-movement procedures, semantic failures are not operationally equivalent: callsign errors can misassign a clearance to the wrong aircraft, taxiway or runway reference errors can produce routing conflicts or unauthorized runway entry, and missed movement constraints can invalidate hold-short, give-way, or boundary restrictions. Because such failure modes correspond to materially different hazard consequences, our evaluation explicitly differentiates entity criticality and action context rather than treating all slot errors uniformly.

Building on this foundation, our evaluation employs a progressive four-level hierarchy: (i) speaker identification, (ii) intention recognition, (iii) risk-aware entity extraction, and (iv) action-level consequence-aware scoring. Unless otherwise stated, all scores are computed on the curated evaluation subset and macro-averaged across classes.

\subsection{Speaker Evaluation}
Speaker role is formulated as a binary classification task over 
\{\textsc{Pilot}, \textsc{Controller}\}. 
We report Macro-F1 as the primary metric and accuracy as auxiliary:
\begin{equation}
\mathrm{Speaker\text{-}F1}=\mathrm{MacroF1}(\textsc{Pilot},\textsc{Controller}).
\end{equation}

\subsection{Intention Evaluation}

Intention classification is modeled as a four-class problem over 
\{\textsc{Greet}, \textsc{Inform}, \textsc{Instruction}, \textsc{Readback}\}. 
Performance is measured using Macro-F1:
\begin{equation}
\mathrm{Intention\text{-}F1}
=\frac{1}{|\mathcal{C}|}\sum_{c\in\mathcal{C}}\mathrm{F1}_c,
\end{equation}
where $\mathcal{C}$ denotes the set of intention classes.

\subsection{Entity Evaluation (Risk-Aware)}
\label{subsec:entity-metric}

\subsubsection{Slot-based representation}
\label{subsubsec:slot-repr}

Entities are evaluated as semantic slots represented by 
\((\textit{entity\_type}, \textit{text})\).

Matching is performed in a one-to-one manner between ground-truth and predicted entities. 
A predicted entity is considered correct if (i) the entity type matches and 
(ii) the token-level overlap between the predicted text and the ground-truth text exceeds a predefined threshold (0.9 in our experiments). 
Each predicted entity can be matched to at most one ground-truth entity.

\subsubsection{Slot-level F1}
\label{subsubsec:slot-f1}

We compute entity extraction performance using Macro-F1 across entity types:
\begin{equation}
\mathrm{MacroF1} = \frac{1}{|\mathcal{E}|} \sum_{e \in \mathcal{E}} \mathrm{F1}_e,
\end{equation}
where $\mathcal{E}$ denotes the set of entity categories. This unweighted formulation ensures that frequent entities do not dominate the overall score.

\subsubsection{Risk-Weighted Entity Recall (RW-ER)}
\label{subsubsec:rwer}
To reflect that not all missing entities carry equal operational impact, we employ a risk-weighted recall over \emph{all ground-truth entities} present in an utterance. Each entity type $e$ is assigned an importance weight $w(e)$ derived from a questionnaire-based survey of five licensed tower controllers from Beijing Capital Airport Tower. Controllers rated the relative safety impact of omitting or misinterpreting each entity category in ground-control communications; ratings were averaged across participants and normalized to $[0,1]$, where higher values indicate greater operational criticality (Tables~\ref{tab:weight-survey} and~\ref{tab:entity-weights}).

Let $\mathcal{E}_{\text{gt}}$ be the set of all ground-truth entities for an utterance and $\mathcal{E}_{\text{hit}}$ be the subset that is correctly predicted. We define:
$$ \mathrm{RW\text{-}ER} = \frac{\sum_{e\in \mathcal{E}_{\text{hit}}} w(e)}{\sum_{e\in \mathcal{E}_{\text{gt}}} w(e)}. $$
This metric penalizes omissions of safety-critical entities (e.g., \textsc{Callsign}, \textsc{Runway/Taxiway}) more heavily than low-impact entities (e.g., \textsc{Greet}).

\begin{table}[t]
\centering
\caption{Summary of expert survey used to derive entity weights.}
\label{tab:weight-survey}
\footnotesize
\setlength{\tabcolsep}{4pt}
\renewcommand{\arraystretch}{1.02}
\begin{tabular}{p{0.28\columnwidth} p{0.62\columnwidth}}
\hline
\textbf{Item} & \textbf{Description} \\
\hline
Format & Questionnaire-based expert rating \\
Participants & 5 licensed tower controllers \\
Source & Beijing Capital Airport Tower \\
Rating target & Relative safety impact of entity categories \\
Aggregation & Mean rating across controllers \\
\hline
\end{tabular}
\end{table}

\begin{table}[t]
\centering
\caption{Entity importance weights $w(e)$ used in risk-aware evaluation. Higher values indicate greater operational criticality if an entity is omitted or misunderstood. O denotes tokens outside annotated entity spans.}
\label{tab:entity-weights}
\footnotesize
\setlength{\tabcolsep}{6pt}
\renewcommand{\arraystretch}{1.05}
\begin{tabular}{l c @{\hspace{10pt}} l c}
\hline
\textbf{Entity} & \textbf{Weight} & \textbf{Entity} & \textbf{Weight}\\
\hline
Callsign   & 1.00 & Taxiway     & 0.90 \\
Runway     & 0.95 & Condition   & 0.95 \\
Vehicle    & 0.65 & Qualifier   & 0.50 \\
Gate       & 0.40 & Report      & 0.40 \\
Frequency  & 0.30 & Controller  & 0.25 \\
Greet      & 0.05 & O (outside) & 0.00 \\
\hline
\end{tabular}
\end{table}

\subsection{Action-Level Risk-Aware Scoring}
\label{subsec:action-risk-metric}

\subsubsection{Action schema and risk levels}
\label{subsubsec:action-schema}
Each utterance is mapped to an action type $a_i \in \mathcal{A}$ with a predefined set of critical slots $\mathcal{S}(a_i)$ as Table~\ref{tab:action-schema}.
Action types are associated with an operational risk level $\rho(a_i)\in\{1.0,0.6,0.2\}$ for \textsc{High}/\textsc{Medium}/\textsc{Low} risk, respectively (determined by domain knowledge and controller feedback).

\begin{table}[t]
\centering
\caption{Action schema with risk levels and critical slots.}
\label{tab:action-schema}
\renewcommand{\arraystretch}{1.05}
\setlength{\tabcolsep}{4pt}
\footnotesize
\begin{tabular}{l l p{0.62\columnwidth}}
\hline
\textbf{Action} & \textbf{Risk} & \textbf{Critical slots $\mathcal{S}(a)$} \\
\hline
HOLD      & High   & callsign, boundary \\
TAXI      & High   & callsign, taxiway, boundary, qualifier, \mbox{runway} \\
GIVE\_WAY & High   & callsign, vehicle \\
CONTACT   & Medium & callsign, frequency, controller \\
PUSHBACK  & Medium & callsign, gate, qualifier \\
INFORM    & Low    & callsign, controller \\
GREET     & Low    & callsign, controller \\
STANDBY   & Low    & callsign \\
UNKNOWN   & Low    & callsign \\
\hline
\end{tabular}
\end{table}

\subsubsection{Risk-weighted action score}
\label{subsubsec:action-score}
Given an utterance $i$ with ground-truth action type $a_i$ and its critical slot set $\mathcal{S}(a_i)$, we define a consequence-aware correctness score:
\begin{equation}
\mathrm{Score}_i
=
r(a_i)\cdot
\frac{\sum_{s\in \mathcal{S}(a_i)} w_{a_i,s}\, m_{i,s}}
{\sum_{s\in \mathcal{S}(a_i)} w_{a_i,s}}.
\label{eq:action-risk-score}
\end{equation}
Here, $w_{a_i,s}$ is the importance weight of critical slot $s$ under action $a_i$ (derived from the entity weights in Table~\ref{tab:entity-weights}), and $m_{i,s}\in\{0,1\}$ is a slot match indicator (1 if slot $s$ is correctly predicted; 0 otherwise).

\paragraph{Action-type risk coefficient.}
We incorporate action-type correctness with a risk coefficient:
\begin{equation}
r(a_i)=
\begin{cases}
1.0, & \text{if action type is predicted correctly},\\
1-\rho(a_i), & \text{otherwise},
\end{cases}
\label{eq:risk-coef}
\end{equation}
where $\rho(a_i)$ is the predefined risk level of action $a_i$:
\begin{equation}
\rho(a_i)=
\begin{cases}
1.0, & \textsc{High risk},\\
0.6, & \textsc{Medium risk},\\
0.2, & \textsc{Low risk}.
\end{cases}
\end{equation}
This design penalizes action-type mistakes more aggressively for high-risk actions (e.g., movement restrictions) while allowing smaller penalties for low-risk exchanges.

\paragraph{Dataset-level aggregation.}
We report the mean action-risk score over all utterances as follows:
\begin{equation}
\mathrm{Action\text{-}RiskScore}=\frac{1}{N}\sum_{i=1}^{N}\mathrm{Score}_i.
\end{equation}

In addition, we report the average inference time per utterance for each model as a practical deployment consideration in real-time operational environments.  

Together, these metrics distinguish surface-level accuracy from true operational understanding, providing a more reliable assessment of deployment readiness under safety constraints.

\section{Experimental Setup}
\label{sec:experiments}

\subsection{Evaluation Data}

All experiments are conducted on the curated evaluation subset described in Section~\ref{sec:dataset}. 
The subset contains 1,000 manually verified ATC utterances with structured annotations, including speaker role, communicative intention, entity slots, and action-level risk labels. 

\subsection{LLM Evaluation Protocol}

We evaluate multiple state-of-the-art commercial and open-weight LLMs via their official APIs under standardized conditions. For each utterance, only the raw text is provided as input.

The prompt specifies the predefined entity types, action types, and a fixed structured output template. Models generate structured outputs accordingly, which are then parsed and evaluated using the risk-aware metrics defined in Section~\ref{sec:metrics}.

No task-specific fine-tuning is applied. All models are evaluated in a strictly zero-shot setting to assess their out-of-the-box reliability. 
This design reflects practical constraints in safety-critical ATC environments, where large-scale, risk-calibrated annotations are scarce and sensitive. 
Evaluating zero-shot behavior establishes a baseline of intrinsic robustness and prevents domain adaptation from masking fundamental structural vulnerabilities.

Inference latency is measured as the average processing time per utterance over the full evaluation set.

\subsection{End-to-End Voice Pipeline}

To evaluate robustness under realistic speech input, we construct an end-to-end voice pipeline by integrating automatic speech recognition (ASR) backbones with downstream semantic parsing. 
Several open-source ASR models (Whisper-large, Whisper-medium, Whisper-small, Whisper-turbo, and wav2vec2) are used to transcribe audio utterances. 
The resulting transcripts are then processed by a fixed downstream LLM (GPT-4o-mini) using the same structured prompting protocol described above. 

Evaluation uses the same risk-aware metrics for direct comparison.
ASR latency is reported separately to reflect practical considerations in real-time operational settings.
\section{Empirical Analysis under Risk-Aware Evaluation}
\label{sec:analysis}

We evaluate state-of-the-art LLMs under the proposed risk-aware metrics to examine whether strong aggregate performance translates into operational safety.

\subsection{Aggregate vs Risk-Aware Gap}

\subsubsection{Overall Performance Landscape}

Under clean transcript input, frontier LLMs show strong structured parsing across speaker, intention, and action extraction. The Gemini series achieves the highest Risk Scores (0.6922 / 0.6884), while GPT-5.1 attains competitive performance (0.6775) and the highest strict score (0.4240). Clear scaling effects are also observed within the Qwen family.

\subsubsection{Conventional Metrics vs.\ Risk-Aware Evaluation}

A consistent gap emerges between conventional metrics (NER F1, Action-Macro) and consequence-aware scores.
For most models, the Risk Score is lower than Action-Macro, indicating that macro averaging overestimates safety performance.

Weighting high-impact entities (e.g., callsigns, runway identifiers, movement constraints) exposes hidden weaknesses.
GPT-5.1 and qwen3-max remain stable under weighting (Act W/T), whereas several other models degrade, revealing sensitivity to critical-slot errors.

\subsubsection{Strict Criterion as a Safety Stress Test}

Under Action-Risk-Strict—where any error in type or critical slot yields zero—performance collapses across all systems.
Most models fall within the 0.1–0.2 range, and even GPT-5.1 (0.4240) remains far from operational reliability. High aggregate semantic accuracy therefore does not imply safety compliance, as minor structural deviations can invalidate an entire instruction under real operational constraints.


\begin{table*}[!ht]
	\centering
	\caption{Model performance comparison with inference latency (1,000 utterances).}
	\label{tab:model-comparison}
	\renewcommand{\arraystretch}{1.05}
	\setlength{\tabcolsep}{3.2pt}
	\footnotesize
	\begin{tabular}{l c c c c c c c c c c}
	\hline
	\textbf{Model} & \textbf{Time (s)} & \textbf{Spk F1} & \textbf{Intent F1} & \textbf{Act F1} & \textbf{Risk-NER} & \textbf{NER F1} & \textbf{Act Macro} & \textbf{Act W/T} & \textbf{Risk Score} & \textbf{Risk Strict} \\
	\hline

	\multicolumn{11}{l}{\textit{Qwen Series}} \\
	qwen3-8b      & 5.19 & 0.3182 & 0.2407 & 0.3367 & 0.4853 & 0.5122 & 0.3962 & 0.3695 & \textbf{0.2880} & 0.1020 \\
	qwen3-14b     & 7.88 & 0.2938 & 0.2103 & 0.4355 & 0.3786 & 0.3914 & 0.3753 & 0.3194 & \textbf{0.2585} & 0.0920 \\
	qwen3-32b     & 9.19 & 0.3987 & 0.2152 & 0.5179 & 0.4021 & 0.4174 & 0.5556 & 0.5586 & \textbf{0.4249} & 0.2060 \\
	qwen-plus     & 4.77 & 0.6550 & 0.4871 & 0.5544 & 0.6236 & 0.6407 & 0.5511 & 0.5717 & \textbf{0.4352} & 0.1640 \\
	qwen3-max     & 7.98 & 0.7003 & 0.5167 & 0.4881 & 0.8293 & 0.8317 & 0.6361 & 0.7348 & \textbf{0.5234} & 0.2440 \\
	\hline

	\multicolumn{11}{l}{\textit{Other Models}} \\
	deepseek-chat & 6.27 & 0.6801 & 0.5699 & 0.6172 & 0.7740 & 0.7801 & 0.5189 & 0.4425 & \textbf{0.4003} & 0.0920 \\
	grok-4-1-fast-reasoning & 20.15 & 0.7252 & 0.6321 & 0.5401 & 0.7793 & 0.7843 & 0.5978 & 0.6062 & \textbf{0.5206} & 0.1809 \\
	\hline

	\multicolumn{11}{l}{\textit{Gemini Series}} \\
	gemini-3-flash-preview & 15.90 & 0.9221 & 0.7865 & 0.6583 & 0.8234 & 0.8236 & 0.7292 & 0.7818 & \textbf{0.6922} & 0.2528 \\
	gemini-3-pro-preview   & 14.60 & 0.9532 & 0.8518 & 0.6423 & 0.8598 & 0.8615 & 0.7265 & 0.7788 & \textbf{0.6884} & 0.2647 \\
	\hline

	\multicolumn{11}{l}{\textit{GPT Series}} \\
	gpt-4o-mini & 3.40 & 0.4111 & 0.2624 & 0.4169 & 0.6120 & 0.6216 & 0.5348 & 0.5443 & \textbf{0.4483} & 0.1630 \\
	gpt-5.1     & 4.00 & 0.7298 & 0.4852 & 0.5478 & 0.7698 & 0.7734 & 0.7512 & 0.8759 & \textbf{0.6775} & 0.4240 \\
	\hline
	\end{tabular}
    
    \vspace{4pt}
    \begin{minipage}{\textwidth}
    \footnotesize
    \textit{Metric abbreviations:}
    Risk Score represents the proposed consequence-aware evaluation metric, where higher values indicate better model performance rather than higher operational risk;
    Act F1 denotes action-type classification F1;
    Risk-NER indicates risk-weighted entity F1;
    NER F1 is unweighted entity-level F1;
    Act Macro refers to unweighted slot-level Macro-F1;
    Act W/T applies slot-importance weighting without type penalty;
    Risk Strict denotes the fail-safe criterion requiring full correctness of action type and critical slots.
    \end{minipage}
    
    \end{table*}


\begin{figure*}[!t]
\centering
\includegraphics[width=\textwidth]{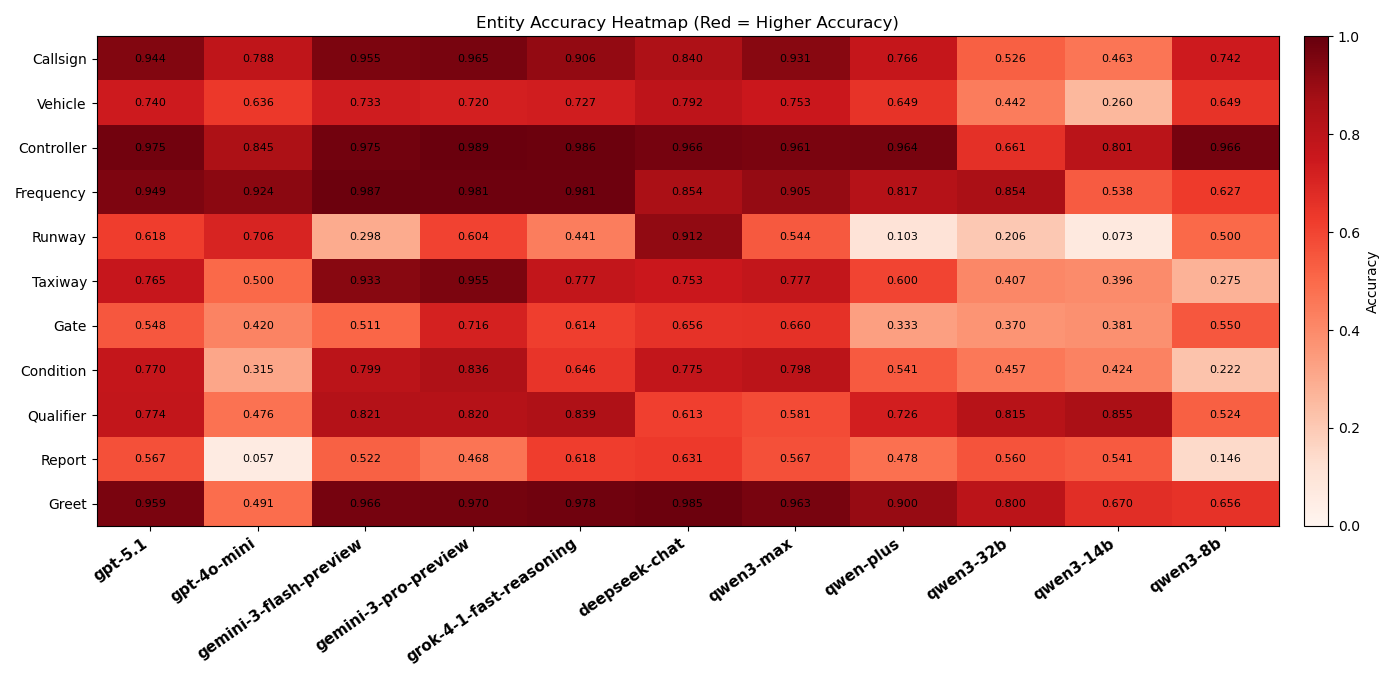}
\vspace{-2mm}
\caption{
Entity-level accuracy across evaluated models. 
Identity-related entities (e.g., \textsc{Callsign}, \textsc{Controller}) achieve consistently high accuracy, 
whereas spatial and constraint-related entities (e.g., \textsc{Runway}, \textsc{Gate}, \textsc{Condition}) exhibit substantial cross-model variability.
}
\label{fig:analysis-picture2}
\vspace{-2mm}
\end{figure*}


\subsection{Structural Failure Patterns}

\subsubsection{Entity-Level Vulnerabilities}

Figure~\ref{fig:analysis-picture2} presents entity-level accuracy across models.
Identity-related entities such as \textsc{Callsign}, \textsc{Controller}, and \textsc{Frequency} exhibit consistently high accuracy across frontier systems, frequently exceeding 0.94.
This suggests that surface-form recognition and role identification are largely saturated under clean transcript conditions.

In contrast, spatial and constraint-related entities demonstrate substantial instability.
\textsc{Runway} extraction remains particularly fragile across model families, with several variants (e.g., qwen-plus, qwen3-14b) dropping below 0.15–0.20 accuracy.
Similarly, \textsc{Gate} and contextual \textsc{Condition} show large cross-model performance gaps, indicating persistent weaknesses in operational grounding and constraint resolution.
These results reveal that aggregate NER F1 obscures safety-critical vulnerabilities concentrated in specific entity categories rather than uniformly distributed errors.


\begin{figure}[!t]
\centering
\includegraphics[width=\columnwidth]{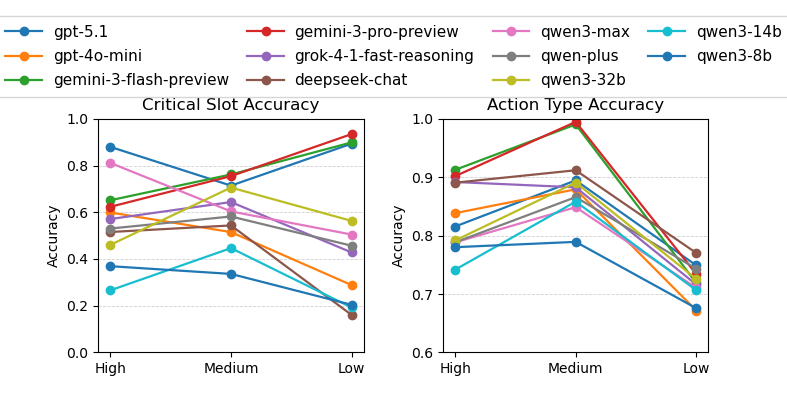}
\caption{
Risk-level robustness curves across evaluated models.
Left: critical slot accuracy; Right: action-type accuracy.
While type performance remains comparatively stable across High, Medium, and Low-risk subsets,
slot extraction exhibits substantially greater variability,
revealing structural grounding instability under safety-critical conditions.
}
\label{fig:risk-level-curve}
\end{figure}

\subsubsection{Risk-Stratified Structural Sensitivity}

To further examine structured failure modes, we conduct risk-stratified evaluation separating action-type classification from critical slot extraction.
Figure~\ref{fig:risk-level-curve} summarizes robustness trends across High, Medium, and Low-risk subsets.

Action-type accuracy remains relatively stable across risk levels, typically fluctuating within a narrow band, and High-risk categories are not consistently the most challenging at the type level. In contrast, slot extraction exhibits higher risk sensitivity. While frontier models (e.g., GPT-5.1 and Gemini variants) maintain relatively stable slot performance across risk strata, several systems show pronounced degradation under high-impact conditions. In multiple cases, models achieve competitive High-risk type accuracy yet display sharp slot-level drops, revealing structural inconsistency in constraint grounding. This type–slot discrepancy explains the sharp decline under strict evaluation: a single error in a safety-critical argument can invalidate an otherwise correct instruction.



\begin{table}[!t]
\centering
\caption{Voice pipeline performance using different ASR models (LLM: GPT-4o-mini).}
\label{tab:voice-comparison}
\renewcommand{\arraystretch}{1.05}
\setlength{\tabcolsep}{2.5pt}
\scriptsize
\begin{tabular}{l c c c c c c c c c c}
\hline
\textbf{ASR} & \textbf{Time} & \textbf{Spk} & \textbf{Int} & \textbf{Act} & \textbf{R-NER} & \textbf{NER} & \textbf{Macro} & \textbf{W/T} & \textbf{Risk} & \textbf{Strict} \\
\hline

\multicolumn{11}{l}{\textit{Clean Transcript}} \\
GPT-4o-mini & -- & 0.41 & 0.26 & 0.42 & 0.61 & 0.62 & 0.53 & 0.54 & \textbf{0.45} & 0.16 \\
\hline

\multicolumn{11}{l}{\textit{Voice Pipeline}} \\
Whisper-L  & 0.83 & 0.33 & 0.25 & 0.38 & 0.18 & 0.21 & 0.24 & 0.08 & \textbf{0.07} & 0.00 \\
Whisper-M  & 0.79 & 0.37 & 0.29 & 0.43 & 0.17 & 0.21 & 0.23 & 0.07 & \textbf{0.07} & 0.00 \\
Whisper-S  & 1.11 & 0.34 & 0.26 & 0.37 & 0.12 & 0.16 & 0.18 & 0.04 & \textbf{0.04} & 0.00 \\
Whisper-T  & 0.34 & 0.33 & 0.26 & 0.42 & 0.15 & 0.18 & 0.23 & 0.07 & \textbf{0.06} & 0.00 \\
wav2vec2   & 0.01 & 0.25 & 0.19 & 0.19 & 0.04 & 0.05 & 0.06 & 0.01 & \textbf{0.01} & 0.00 \\
\hline
\end{tabular}
\end{table}

\subsection{Deployment Fragility under ASR Noise}

\subsubsection{Voice Pipeline Degradation}

Table~\ref{tab:voice-comparison} reports performance under the end-to-end voice pipeline.
Introducing ASR noise causes a sharp degradation across all consequence-aware metrics. While GPT-4o-mini achieves a Risk Score of 0.45 on clean transcripts, performance drops to 0.04--0.07 with Whisper backbones and further to 0.01 with wav2vec2, representing an order-of-magnitude decline despite identical downstream inference. Action-Macro decreases more moderately than risk-weighted metrics, indicating that coarse action-type recognition remains partially intact whereas structured slot extraction—particularly callsigns, runway identifiers, taxiway references, and boundary constraints—is highly sensitive to even minor transcription perturbations. As a result, errors that appear minor at the lexical level propagate directly to consequence-aware scoring.

\subsubsection{Safety Collapse under Strict Criterion}

The impact becomes critical under the fail-safe Action-Risk-Strict metric.
Once strict structural correctness is required, performance collapses to 0.00 across all ASR configurations.
No voice-based setting produces fully correct structured predictions when transcription noise is present.

These findings reveal a substantial gap between transcript-level evaluation and realistic deployment conditions.
Even small recognition inconsistencies can invalidate safety-critical instructions under operational constraints.
Text-only benchmarking therefore substantially overestimates real-world reliability and should be complemented by end-to-end voice evaluation.

\section{Conclusion}
\label{sec:conclusion}

This work evaluates large language models for safety-critical language understanding in air traffic control (ATC) under consequence-aware metrics.
We show that conventional aggregate metrics substantially overestimate operational reliability.
Under risk-weighted scoring and strict structural correctness, performance drops markedly across all models.

Failures concentrate in safety-critical entities such as runway identifiers and movement constraints, even when action types are predicted correctly.
Deployment under realistic voice conditions further induces an order-of-magnitude degradation, with strict correctness collapsing across ASR settings.
These results indicate that high-level semantic accuracy does not guarantee structured operational safety.

Future work will investigate safety-oriented model improvement, constraint-aware decoding, domain adaptation, more extensive sensitivity analysis, and human-in-the-loop validation, together with broader cross-domain extension of the evaluation setting. More broadly, evaluation frameworks for high-risk domains must move beyond linear token-level averages toward consequence-sensitive assessment.

Reliable deployment in safety-critical environments requires not only stronger models, but stricter evaluation standards.




	\bibliographystyle{IEEEtran}
	\bibliography{root} 
	
\end{document}